\pdfoutput=1

\documentclass[11pt]{article}

\usepackage[final]{acl}

\usepackage{times}
\usepackage{latexsym}

\usepackage[T1]{fontenc}

\usepackage[utf8]{inputenc}

\usepackage{microtype}

\usepackage{inconsolata}

\usepackage{graphicx}

\usepackage{amsmath}
\usepackage{subcaption}

\title{Dissecting Persona-Driven Reasoning in Language Models \\via Activation Patching}

\author{
 \textbf{Ansh Poonia\textsuperscript{$\dag$}},
 \textbf{Maeghal Jain\textsuperscript{$\dag$}}
\\
 \textsuperscript{$\dag$}Independent
\\
\texttt{
    \{\href{mailto:pooniaansh11@gmail.com}{pooniaansh11}, \href{mailto:maeghaljain@gmail.com}{maeghaljain}\}@gmail.com
  }
\\
}

\begin{document}
\maketitle
\begin{abstract}
Large language models (LLMs) exhibit remarkable versatility in adopting diverse personas. In this study, we examine how assigning a persona influences a model's reasoning on an objective task. Using activation patching, we take a first step toward understanding how key components of the model encode persona-specific information. Our findings reveal that the early Multi-Layer Perceptron (MLP) layers attend not only to the syntactic structure of the input but also process its semantic content. These layers transform persona tokens into richer representations, which are then used by the middle Multi-Head Attention (MHA) layers to shape the model's output. Additionally, we identify specific attention heads that disproportionately attend to racial and color-based identities.\footnote{Code and some additional results are available at \href{https://github.com/anshpoonia/Persona-Driven-Reasoning}{https://github.com/anshpoonia/Persona-Driven-Reasoning}}
\end{abstract}

\section{Introduction}
Recent advances in large language models (LLMs) have demonstrated their striking ability to adopt a wide range of personas, enabling context-sensitive and tailored responses (\citealp{serapio2023personality}, \citealp{zhang2024personalization}, \citealp{joshi-etal-2024-personas}, \citealp{sun-etal-2025-persona}). However, studies such as those by \citealp{salewski2023incontext}, \citealp{deshpande-etal-2023-toxicity}, \citealp{gupta2024bias}, \citealp{zheng-etal-2024-helpful} show that persona assignment can significantly influence reasoning and, in some cases, amplify underlying social biases. While these works focus on identifying and quantifying such effects, they do not examine the causal mechanisms within a pre-trained language model (PLM) that give rise to them.

Mechanistic interpretability provides framework for uncovering the causal mechanisms by which language models carry out specific tasks. \citet{lieberum2023does} introduced this approach in the context of multiple-choice question answering (MCQA), identifying "correct letter heads" in a 70B Chinchilla model-attention heads that track answer symbols and promote the correct choice based on positional order. \citet{wiegreffe2025answerassembleaceunderstanding} examined how successful models implement symbol binding in formatted MCQA using vocabulary projection and activation patching, and \citet{li2025anchoredanswersunravellingpositional} analyzed anchored positional bias in the GPT-2 family, showing that models consistently favor the first option ('A'). Building on these interpretability techniques, we ask whether similar internal circuits also govern how personas steer a model’s reasoning and potentially introduce social bias.

This study takes a step toward bridging the gap between surface-level observations of persona effects and the underlying mechanisms that produce them. We investigate the roles of key model components namely, Multi-Layer Perceptron (MLP) layers, Multi-Head Attention (MHA) layers, and individual attention heads in shaping the reasoning shifts induced by persona assignments. Using activation patching, we probe the internal circuitry of LLMs to trace the origins of persona-driven variation in objective tasks. Our findings challenge the prevailing assumption that early MLP layers are concerned solely with syntactic processing. We show that these layers also encode semantic features related to persona. Furthermore, we identify a small number of attention heads that disproportionately focus on tokens associated with race-based persona cues. Although our work is limited to uncovering the origin of persona-driven behavior, it contributes to a deeper understanding of LLMs and lays groundwork for future efforts to mitigate deep-seated biases in these systems.

\section{Experimental Setup}

\subsection{Dataset Details}
We chose the MMLU dataset \citep{hendrycks2021measuring} for this study, which contains  14,024 multiple-choice questions spanning 57 subjects. We selected this dataset for two main reasons. First, the objective, multiple-choice format allowed us to focus on a fixed set of four answer tokens, giving us a well-defined target. Second, the wide range of subject areas helped reduce domain-specific or persona-driven biases. The identities or personas\footnote{The terms persona and identity have been used interchangeably in this work.} we examined in our study fall into four broad categories: racial identities, color-based identities, and identities defined by positive or negative attributes; details of each can be found in the Appendix \ref{sec:identities}. To maintain consistency and minimize variance, we used \textit{student} as a gender-neutral subject across all defined identities. For instance: \textit{Asian student}, \textit{white student}, \textit{good student}, etc.

\subsection{Model and Prompt}
Our primary investigation was conducted using the Llama 3.2 1B Instruct \citep{llama} model to accommodate our computational constraints. We supplement these main findings with additional results from experiments on the Llama 3.2 3B Instruct \citep{llama} and Qwen 2.5 1.5B Instruct \citep{qwen2025qwen25technicalreport}. Compact size of these models makes it well-suited for efficient experimentation, while still delivering strong performance relative to other open-source models in their class. Our methods are scalable and can be extended to larger models with sufficient computational resources and minor adjustments. The Instruct variant of the models also permits the use of system prompts\footnote{System prompts are instructions given to the AI before any user input. They define the AI's behavior, role, and response style throughout an interaction.}, allowing us to specify the identity the model should adopt in its responses. A standard system prompt serves as the base for each question, enabling identity shifts by modifying only two tokens in the entire prompt: just one token distinguishes each identity. We used the prompt structure defined by Meta and Qwen team for MMLU dataset \citep{llama-prompt}, with the addition of the system prompt, see Appendix \ref{sec:prompt}.

\section{Persona Evaluation}
We computed model outputs using zero-shot prompts to isolate the variation introduced solely by changes in the persona token, avoiding any influence from patterns introduced by few-shot prompting. This also reduced the length of each prompt, allowing for faster computation. For each prompt, we calculated the probability of the next token, which, by design, corresponds to the selected answer for the given question. This process was repeated for all 16 identities, and also the base identity.

\begin{figure}
    \centering
    \includegraphics[width=\linewidth]{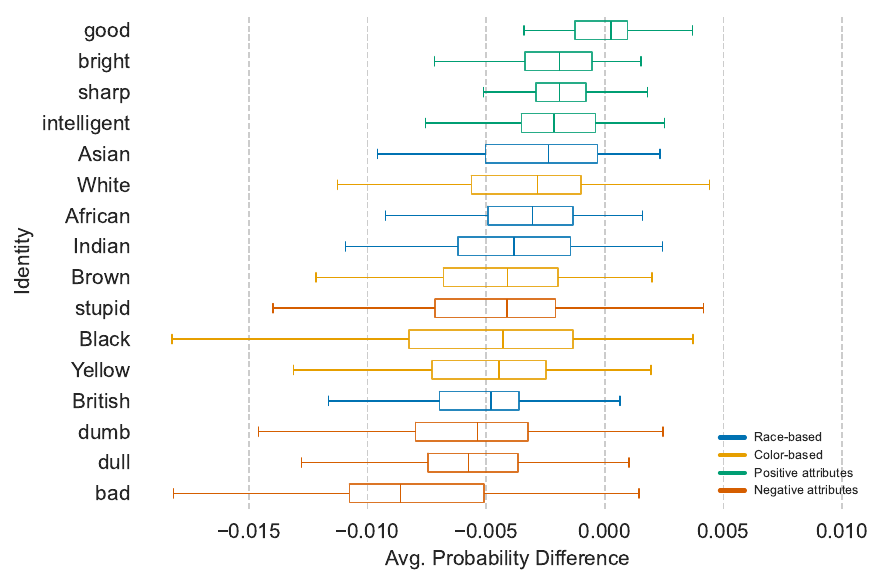}
    \caption{Average difference in the probability of the correct token for each identity, relative to the base prompt.}
    \label{fig:prob_diff}
\end{figure}

Our main focus is the change in the probability of the correct token for each persona relative to the base identity, as shown in Figure~\ref{fig:prob_diff}. This highlights shifts in reasoning attributable to persona alone. In some cases, the differences in probability followed patterns that appeared to have a semantic basis. For instance, the negatively attributed student persona performed significantly worse than those described with positive attributes, having an average probability difference of -0.0027 (T = 11.7, p < 0.001). In contrast, the results for racially or color-coded personas were less consistent, and no definitive conclusion could be drawn about whether the patterns reflect stereotypical associations. A similar trend was observed when examining probability differences across identities for other models, see Appendix \ref{sec:prob_diff} for details. Overall, our results suggest that the model's reasoning ability varies even with minimal changes to the persona being imitated.

\begin{figure*}
    \centering
    \begin{subfigure}[t]{0.49\linewidth}
        \centering
        \includegraphics[width=\linewidth]{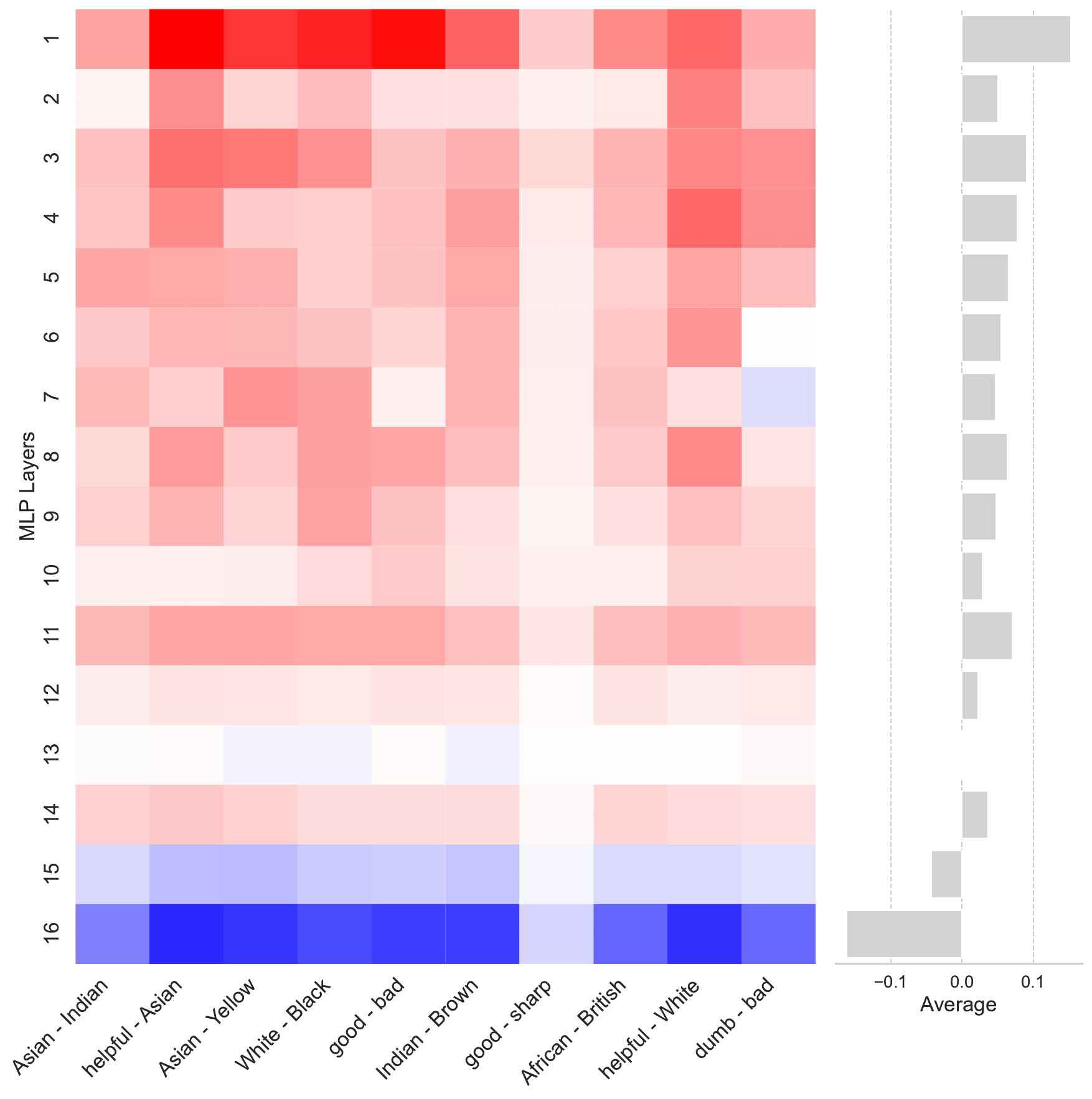}
        \caption*{MLP layer patching}
    \end{subfigure}
    \hfill
    \begin{subfigure}[t]{0.49\linewidth}
        \centering
        \includegraphics[width=\linewidth]{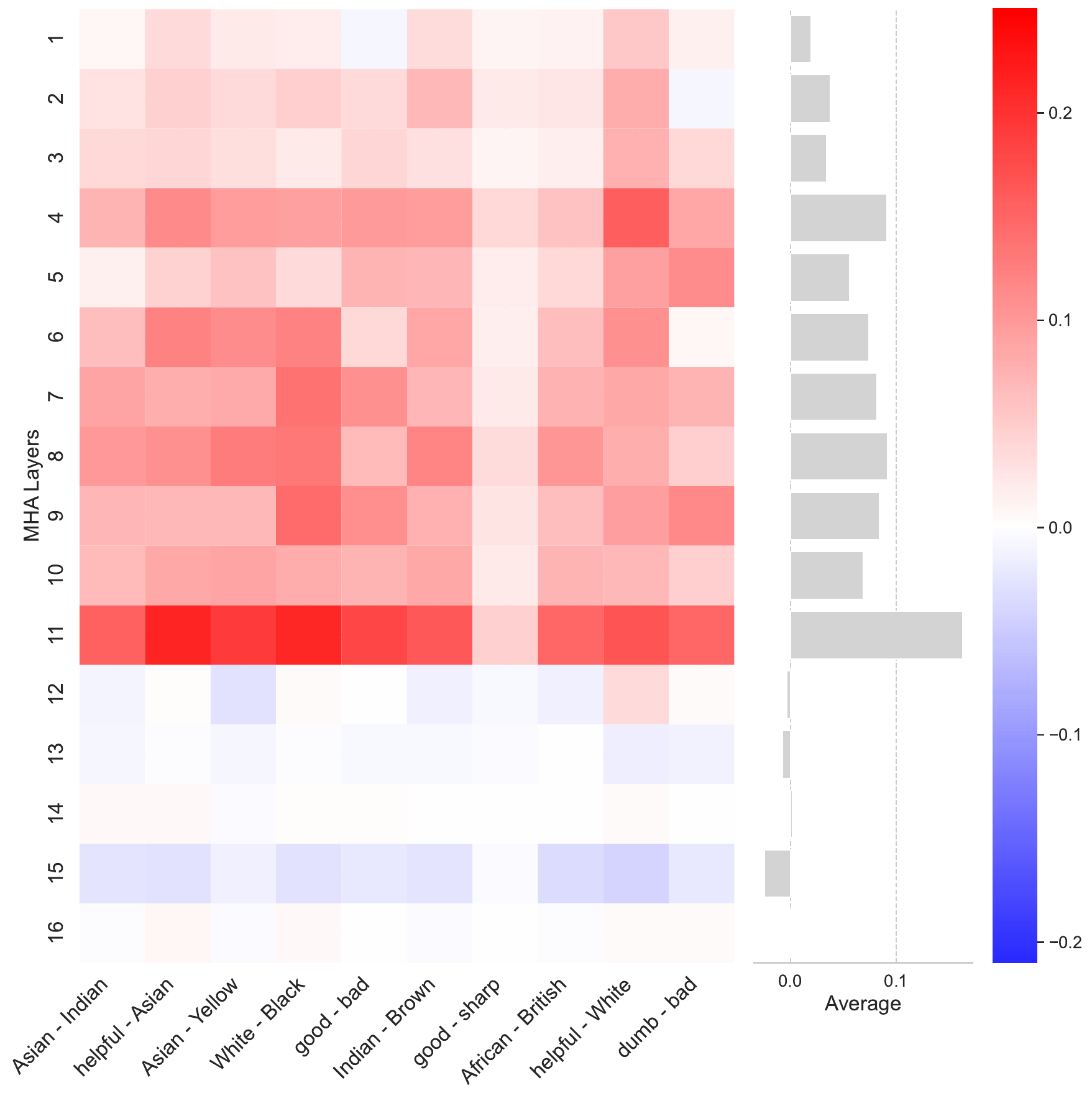}
        \caption*{MHA layer patching}
    \end{subfigure}
    \caption{Relative logit difference ($\Delta_r$) when MLP layer (left) and MHA layer (right) is patched in Llama 1B model. Accompanying bar chart shows the average $\Delta_r$ across identity pairs.}
    \label{fig:patching_results}
\end{figure*}

\section{Interpretability Investigation}
\subsection{Methods}
We focus on activation patching, also referred to as Resample Ablation, or Causal Mediation Analysis (\citealp{Vig2020}, \citealp{meng2022locating}), to understand the influence of individual components on the model's selection of the correct answer. Specifically, we employ "de-noising" variant of activation patching, where the activation of a component from a corrupted run is replaced with the corresponding activation from a clean run \citep{heimersheim2024use}. The clean and corrupted inputs are constructed using Symmetric Token Replacement, a technique in which only one or a few tokens are altered (\citealp{meng2022locating}, \citealp{wang2023interpretability}). This approach doesn't throw the model's internal state out-of-distribution and preserves the syntactic structure of the input \citep{zhang2024towards}. For each experiment, we select a pair of personas to generate the clean \texttt{(ID1)} and corrupted \texttt{(ID2)} prompts. A total of ten persona pairs were chosen keeping computational constraints in mind; details are provided in the Appendix \ref{sec:id_pairs}. Care was taken to balance pairings identities both within and across categories to ensure reliable results. Pairings with a base identity were also included for comparison. For each pair, we divide the full set of questions into four subsets, Appendix \ref{sec:subsets}. From these, we select the subset in which the first persona \texttt{(ID1)} answers correctly while the second persona \texttt{(ID2)} answers incorrectly. In each such pair of prompts, the only difference lies in the token representing the persona. As a result, any change in the logit of the correct answer after patching reflects the influence of the component (and its downstream effects) on how the model processes persona-related information. 

\subsection{Effects}
The impact of patching on the model's output can be broken down into direct and indirect effects \citep{pearl2001}. The direct effect measures the isolated contribution of the component to the output. The indirect effect captures the influence the component has via the behavior of later layers and is computed by subtracting the direct effect from the total effect of patching the component. The direct effect of a given component can be measured by subtracting its output from the residual stream before the final-layer norm in the \textit{corrupt run}, and adding the output of that component from the \textit{clean run}. This preserves the indirect effect of the component and only alters its direct effect. Together, these effects help characterize the role the component plays in the processing of input information.

\subsection{Metrics}
To quantify the impact of patching, we use two metrics. First, we check whether patching causes the logit ($l$) of the correct answer to become the highest among all four options. Second, we measure the relative logit difference ($\Delta_{r}$), i.e., change in the logit of the correct answer relative to the change in the mean logit across all options.
\begin{multline*}
    \Delta_{r} = \{l_{correct}(P) - l_{correct}(C)\} \\
    - \{\mu(l_{ABCD}(P)) - \mu(l_{ABCD}(C))\}
\end{multline*}

Here, $l_{*}(P)$ is the logit from patched run and $l_{*}(C)$ from corrupt run, and $\mu$ is mean function. We use this relative metric rather than an absolute one because some components may support the correct answer not by increasing its logit directly, but by decreasing the logits of incorrect options. An absolute measure, like simple logit difference, would overlook such cases and only highlight components with large magnitude effects on the correct answer. These metrics are computed across all selected questions for each persona pair.

\subsection{Findings}
We begin by patching the MLP and MHA layers across all token positions. The Figure~\ref{fig:patching_results} shows $\Delta_{r}$ across all identity pairs, see Appendix \ref{sec:additional} for results on other metric, and Appendix \ref{sec:patch_other} for patching results of other two models. Initial observations can be made by comparing the semantic similarities of identity pairs with the results from activation patching. The \textit{good-sharp} identity pair, for instance, produced a noticeably lower $\Delta_{r}$ score across all models. This is likely because the terms are semantically similar; a \textit{good student} and a \textit{sharp student} both represent a capable student, and the model gives nearly identical probabilities to the correct answer for each. A similar, though less distinct, pattern can be seen with the \textit{bad-dumb} identity pair. When identities like \textit{Asian} and \textit{White} are patched with activations from the neutral base identity, the effect on the model's reasoning is much higher. Since the base identity is a neutral persona, the fact that patching activations from it improves the model's performance suggests that these components are introducing bias in the presence of specific identities. This bias hampers the model's reasoning, and patching the output of these components essentially prevents this bias from occurring. The lower $\Delta_{r}$ values for pairs like \textit{White-Black} in Llama 3B and \textit{Asian-Yellow} in Qwen 1.5B may be a result of the specific ways the models were fine-tuned, as this inconsistency was not observed in other models.

\begin{figure}
    \centering
    \includegraphics[width=\linewidth]{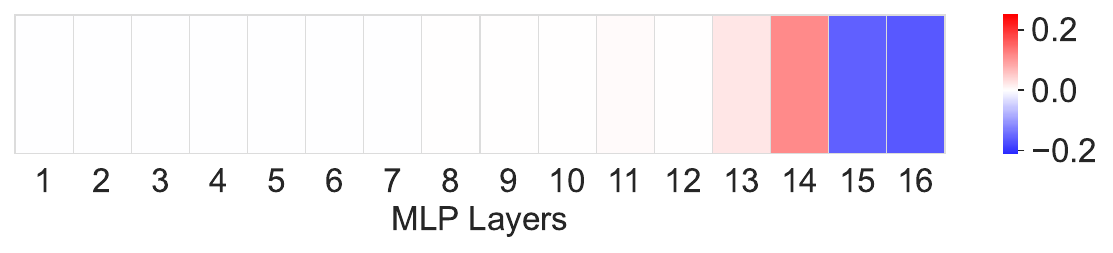}
    \caption{Average $\Delta_r$ when only direct component of a MLP layer is patched.}
    \label{fig:mlp_direct}
\end{figure}

We observe that patching the early MLP layers (layers 1–3) and the middle MHA layers (layers 9–11) produces a consistently higher total effect across all ten identity pairs. Similar observations can be made for Llama 3B model, but in additions to these, 18th MLP layer and 1st MHA in Qwen also gives higher $\Delta_{r}$, which requires further exploration in future studies. We hypothesize that \textit{the early MLP layers develop persona-specific information, particularly at the token position(s) representing the identity. This information is then picked up and used by the later MHA layers, giving rise to the observed persona-driven behavior.} To test this, we patch only the activations at the identity token position  in the MLP layers, rather than all positions. We observed that patching activations at the identity token position in the first MLP layer produced an effect nearly equivalent to patching all token positions. Beyond the first layer, the effect rapidly decays, with later layers showing little to no impact, see Appendix \ref{sec:id_patch}. 

We also find that only the MLP layers near the end of the network have any direct effect on the output, see Figure~\ref{fig:mlp_direct}. This implies that the effect seen in the initial layers is purely indirect. Since the local syntax around the point of change remains constant, this suggests that the initial MLP layers are processing not just structure but also the semantics of the input tokens. This counter to earlier findings (\citealp{belinkov-etal-2017-evaluating}, \citealp{peters-etal-2018-dissecting}, \citealp{jawahar-etal-2019-bert}) that claimed these layers were focused only on syntactic or local features, leaving semantic processing to later layers. The subtlety of this semantic processing may explain why it was overlooked in previous studies. Our results align with observations that initial MLP layers transform tokens into richer representations \citep{stolfo-etal-2023-mechanistic}, supporting the hypothesis that these layers induce persona-specific semantics that later layers utilize.

\begin{figure}
    \centering
    \includegraphics[width=\linewidth]{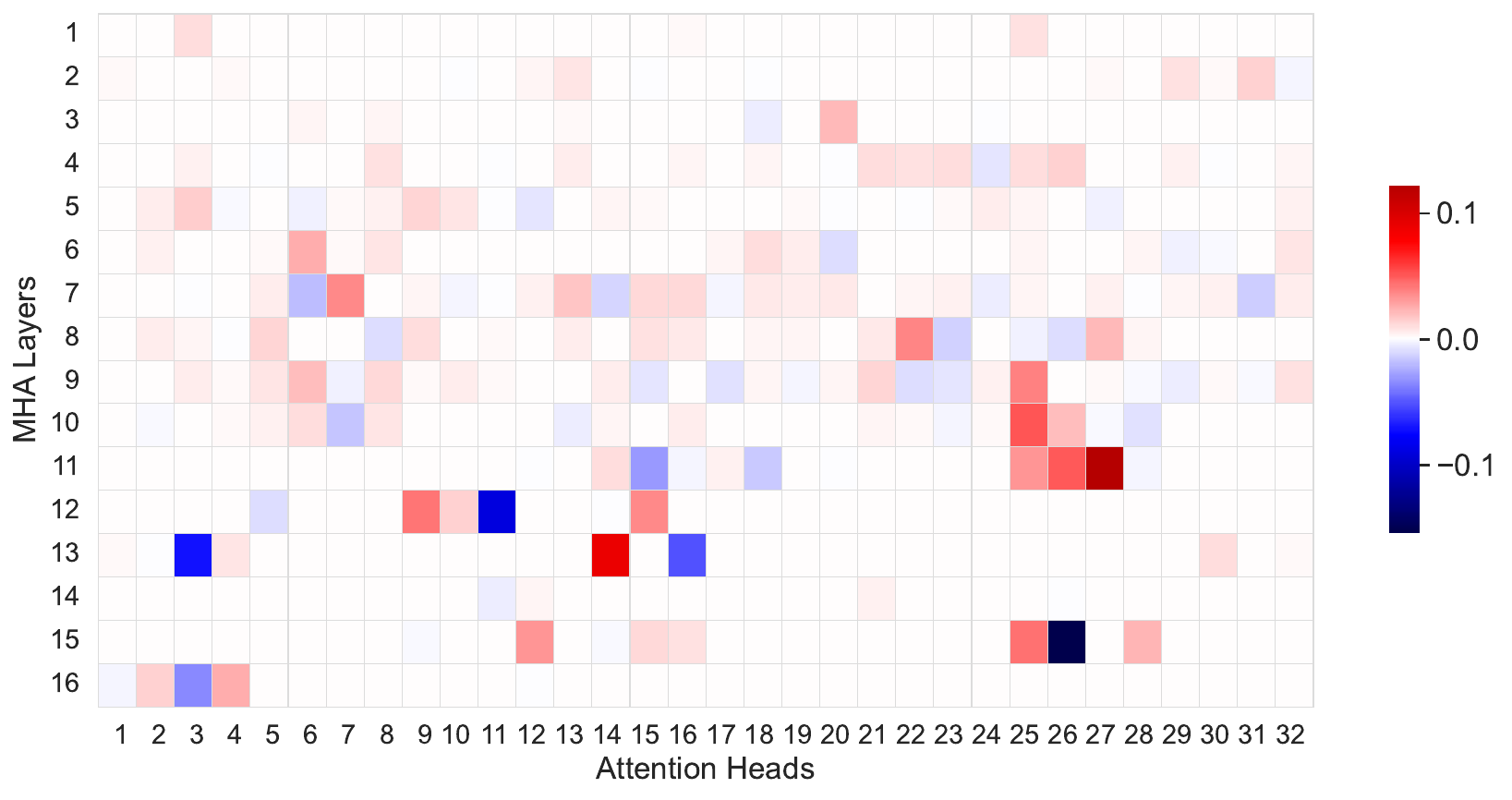}
    \caption{Average $\Delta_r$ for patching individual attention heads.}
    \label{fig:attention_heads}
\end{figure}

To investigate MHA layer roles more closely, we performed activation patching on individual attention heads ($H_{layer}^{head\ number}$), identifying eight heads with a high positive effect on the output and four with a high negative effect, see Figure~\ref{fig:attention_heads}. We analyzed the value-weighted attention patterns \citep{lieberum2023does} of these heads on the identity token position across five questions per subject, categorizing them based on the relative attention given when compared with other identities, see Appendix \ref{sec:attn_vis}. $H_9^{25}$, $H_{10}^{25}$, $H_{11}^{26}$, $H_{13}^{3}$, and $H_{13}^{14}$ consistently allocated higher attention to tokens representing racial identities. $H_{11}^{26}$, $H_{13}^{3}$, and $H_{13}^{14}$ also showed elevated attention to color-based identities, though this pattern was less consistent across domains. $H_{13}^{16}$ uniquely focused on color-based identities, while $H_{15}^{25}$ prioritized both positive and negative attributed identities at early token positions. $H_9^{25}$ also gives negative identities more attention at token positions near the end. We further examined how these attention patterns of these heads responded to MLP layer patching. When activations from runs with racial or color-based personas were replaced with those from positive or negative attributed personas, the attention of heads previously showing high focus on the identity token position decreased significantly. This reduction occurred regardless of whether all token positions or only the identity token position were patched. For most heads, patching layers beyond the first had minimal impact on attention patterns, for details see Appendix \ref{sec:attn_patch}. 

Our findings validate the hypothesis of an interaction between the initial MLP layers and the middle MHA layers in driving persona-driven behavior. The initial MLP layers, especially at the identity token position, form rich, persona-specific semantic representations. These are then taken up by the middle MHA layers, impacting the model's choice of response.

\section{Discussion}
In this work, we analyzed the impact of persona-driven behavior on the reasoning ability of a language model in objective tasks. Through the lens of mechanistic interpretability, we examined the role of different component models: MLP layers, MHA layers, and individual attention heads in shaping this behavior. We observed that early MLP layers also contributes towards semantic understanding of inputs and encode persona-specific information into richer representation. This information is then passed to later MHA layers, which use it to influence the model's response. We further categorized attention heads based on their relative attention patterns and identified a subset that assigned disproportionate weight to racial and color-related attributes associated with a persona.

These findings also have significant practical implications for developing fairer and more reliable AI systems. The research demonstrates how to pinpoint specific model components, like certain MLP layers and attention heads, that encode and act upon persona-driven biases related to race and other attributes. This allows for interventions that are far more targeted than standard fine-tuning. Instead of making broad adjustments, we can directly modify, steer, or even disable the specific neural circuits responsible for undesirable stereotypical behaviors, leading to more effective and efficient bias mitigation. Furthermore, when coupled with methods like probing, it serves as a powerful diagnostic tool, enabling cheaper, more precise monitoring of how a model processes sensitive information and offering a clear window into its internal reasoning. This deeper interpretability is essential for debugging, ensuring AI safety, and building systems that are not only less biased but also more transparent and controllable in how they adopt these biases.

Overall, our observations offer preliminary insight into the subtle yet significant functions of certain model components and how they can be revealed through constrained but straightforward experiments. We took initial steps toward understanding the origins of persona-driven behavior in LLMs. In future studies, we will investigate why certain personas answer specific questions correctly while others do not, and how output vector circuits in attention heads use earlier-layer representations to shape final predictions. These directions will lead to a deeper understanding of what "persona" truly means in the context of language models.

\section*{Limitations}
Although the MMLU dataset contains a large number of mostly factual, objective questions, we acknowledge that it is not the only dataset that meets our selection criteria. Our experiments were conducted on the Llama 3.2 1B, Llama 3.2 3B and Qwen 2.5 1.5B Instruct models, while attention heads level patching experiments being limited to Llama 1B model, due to computational and time constraints. However, the methods described here can be readily extended to models of different sizes and architectures, as well as to other datasets with similar characteristics.

We selected a set of 16 personas to approximate the space relevant to our probing efforts. The list of personas is not exhaustive, but it serves as a practical starting point. Our analysis focused on attention heads identified as important through activation patching. Examining additional heads may improve our understanding of how persona-driven behavior develops within the model. At present, however, attention pattern analysis requires manual inspection, which remains a slow and labor-intensive process.

\bibliography{main}

\appendix

\section{Identities}
\label{sec:identities}

In this study, personas or identities refer to single words such as "Asian" or "good." The selected identities fall into four broad categories, with four identities chosen from each to ensure balanced comparisons. A further criterion in selection was that each identity should be of a single token, so that the only variation between prompts would be one token. Table~\ref{tab:identities} lists the identities alongside their respective categories.

\begin{table}[]
    \centering
    \begin{tabular}{ll}
        \hline
        \textbf{Category} & \textbf{Identity} \\
        \hline
        Racial-based & Asian \\
        & Indian \\
        & African \\
        & British \\
        \hline
        Color-based & White \\
        & Black \\
        & Brown \\
        & Yellow \\
        \hline
        Positive-attributes & good \\
        & intelligent \\
        & bright \\
        & sharp \\
        \hline
        Negative-attributes & bad \\
        & dull \\
        & stupid \\
        & dumb \\
        \hline
    \end{tabular}
    \caption{Identities with their respective category.}
    \label{tab:identities}
\end{table}

\begin{table}[]
    \centering
    \begin{tabular}{c|c}
        \hline
        \textbf{ID1} & \textbf{ID2} \\
        \hline
        Asian student & Indian student \\
        helpful assistant & Asian student \\
        Asian student & Yellow student \\
        White student & Black student \\
        good student & bad student \\
        Indian student & Brown student \\
        good student & sharp student \\
        African student & British student \\
        helpful assistant & White student \\
        dumb student & bad student \\ 
        \hline
    \end{tabular}
    \caption{Identity Pairs}
    \label{tab:id_pairs}
\end{table}

\begin{figure}
    \centering
    \includegraphics[width=0.95\linewidth]{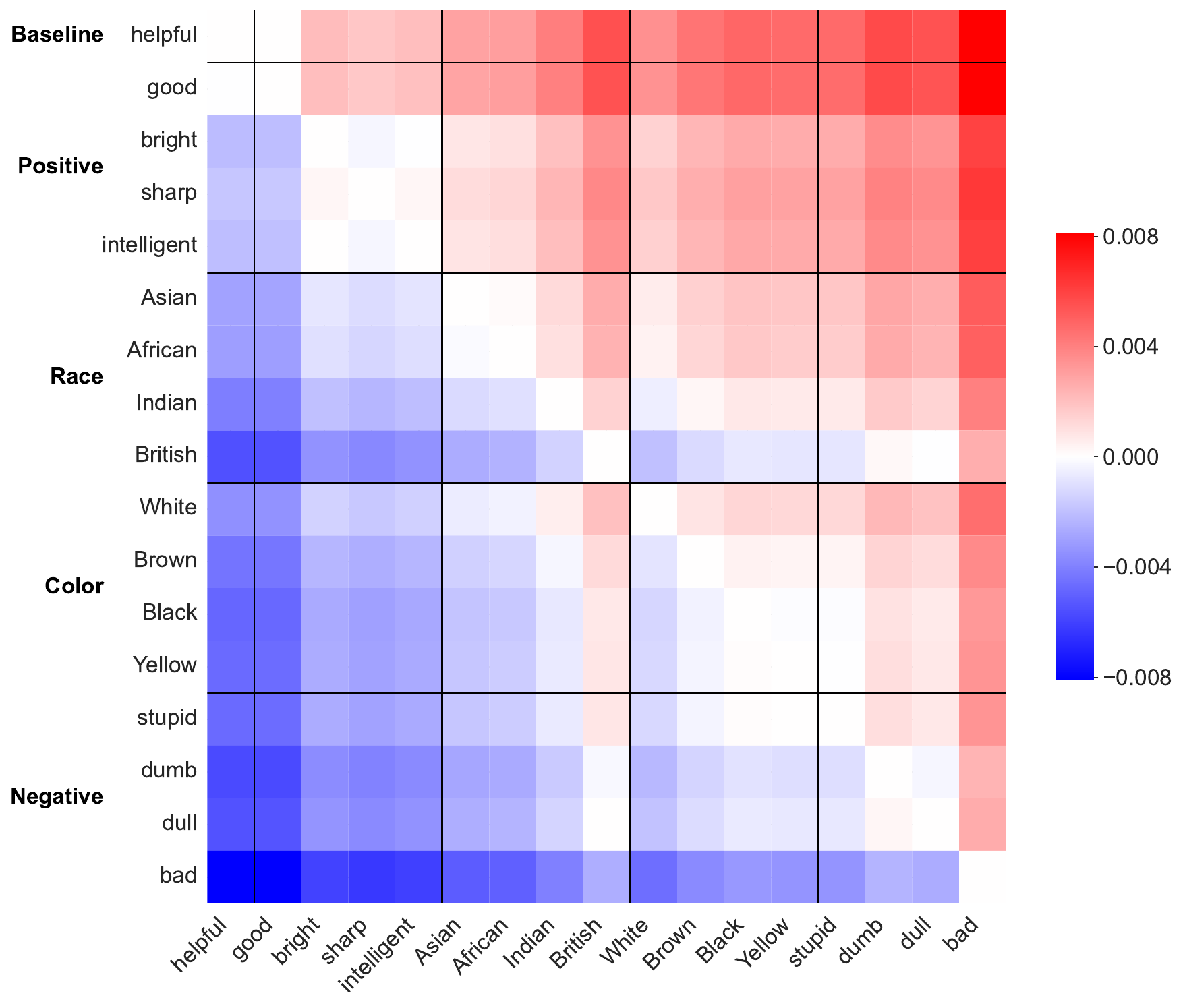}
    \caption{Average difference in probability of correct token of identities w.r.t. each-other.}
    \label{fig:prob_change_ids}
\end{figure}

\section{Prompt Structure}
\label{sec:prompt}

\begin{figure*}[t]
    \centering
    \begin{subfigure}{\linewidth}
        \centering
        \includegraphics[width=0.8\linewidth]{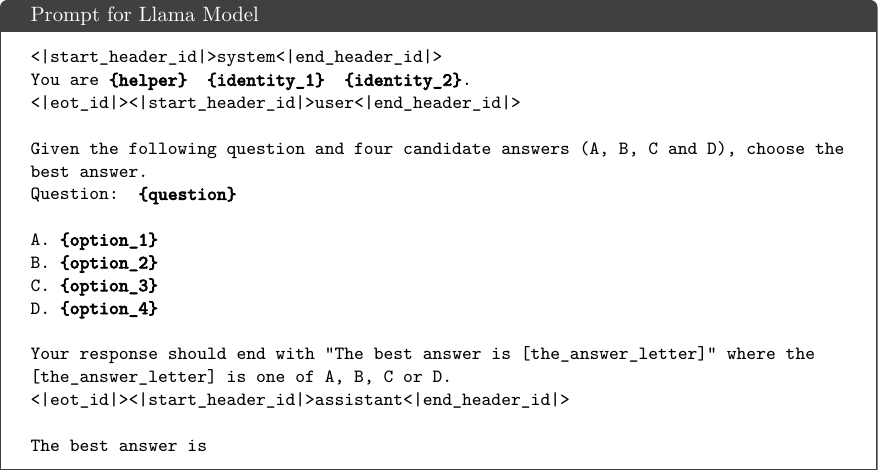}
    \end{subfigure}

    \vspace{-0.4em}

    \begin{subfigure}{\linewidth}
        \centering
        \includegraphics[width=0.8\linewidth]{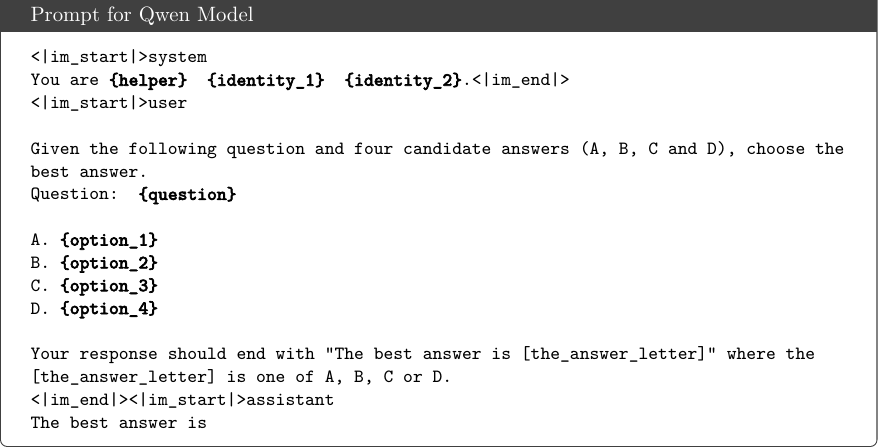}
    \end{subfigure}

    \caption{MMLU prompt structure.}
    \label{fig:prompt_designs}
\end{figure*}

The prompt format used throughout the study is shown in Figure~\ref{fig:prompt_designs}. The placeholder \texttt{\{helper\}} is replaced with "a" or "an" depending on whether the first letter of \texttt{\{identity\_1\}} is a vowel. The variable \texttt{\{identity\_1\}} is substituted with the identities listed in Table~\ref{tab:identities}, or with "helpful" in the base prompt. The placeholder \texttt{\{identity\_2\}} is replaced with "student" in the persona prompts and with "assistant" in the base prompt. The \texttt{\{question\}} field is filled with the target question, and each \texttt{\{option\_*\}} is replaced by the corresponding answer choice.

\section{Additional Performance Results}
\label{sec:prob_diff}
\begin{figure*}
    \begin{subfigure}[t]{0.49\linewidth}
        \centering
        \includegraphics[width=\linewidth]{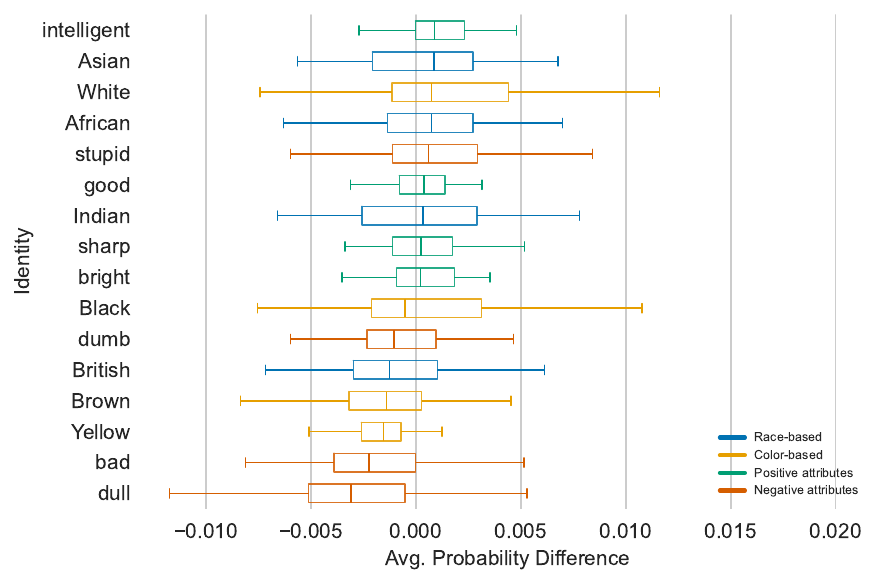}
        \caption*{Llama 3B}
    \end{subfigure}
    \hfill
    \begin{subfigure}[t]{0.49\linewidth}
        \centering
        \includegraphics[width=\linewidth]{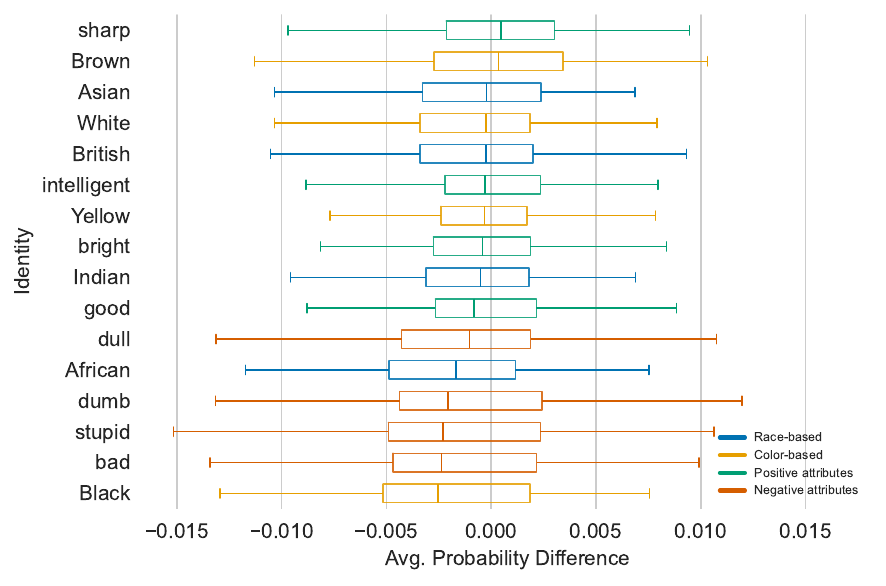}
        \caption*{Qwen 1.5B}
    \end{subfigure}

    \caption{Results for Llama 3B (left) and Qwen 1.5B (right) for average difference in the probability of correct tokens w.r.t baseline.}
    \label{fig:accuracy_diff}
\end{figure*}

The average of relative difference in the probability assigned to correct token for each identity w.r.t. base identity, for Llama 3B and Qwen 1.5B is shown in Figure~\ref{fig:accuracy_diff}. For Llama 1B model, average of difference in the probability assigned to correct token for a given identity relative to other identities is shown in Figure~\ref{fig:prob_change_ids}.

\section{Identity Pairs}
\label{sec:id_pairs}

Table~\ref{tab:id_pairs} shows the selected identity pairs. The pairing of identity terms was partly based on common stereotypes, such as \textit{Asian-Yellow}, \textit{Indian-Brown}, \textit{Asian-Indian}, and contrasted with pairs like \textit{White-Black} and \textit{African-British}. Other identities were chosen from sets of positive and negative attributes, to include both semantically similar pairs (\textit{good-sharp}, \textit{dumb-bad}) and dissimilar ones (\textit{good-bad}). A few were added for comparison with baseline identities such as \textit{helpful-Asian} and \textit{helpful-White}. During activation patching, the prompt for \texttt{ID1} serves as the clean prompt, while the prompt for \texttt{ID2} serves as the corrupt prompt. Activations from the \texttt{ID2} run are replaced with those from the \texttt{ID1} run to identify which components restore the model's output to that of \texttt{ID1}. The only difference between the \texttt{ID1} and \texttt{ID2} prompts is the identity token, except when \texttt{ID1} corresponds to the base prompt, in which case \textit{assistant} is replaced with \textit{student}. 

\section{Question Subsets}
\label{sec:subsets}

For each identity pair \texttt{(ID1, ID2)}, the questions in the dataset are divided into four subsets: \texttt{S1} – questions answered correctly by both identities; \texttt{S2} – questions answered incorrectly by both; \texttt{S3} – questions answered correctly by \texttt{ID1} but incorrectly by \texttt{ID2}; \texttt{S4} – questions answered correctly by \texttt{ID2} but incorrectly by \texttt{ID1}. The number of questions in each category for the selected identity pairs is shown in Table~\ref{tab:subsets}.

Questions from subset \texttt{S3} were chosen for activation patching because they offer a well-defined target token to observe during the patched run. In the \texttt{ID2}\textit{(corrupt)} run, the logit of the correct token is lower than in the \texttt{ID1}\textit{(clean)} run. Therefore, components that raise the logit of the correct token during the patched run help restore the model's behavior to that of \texttt{ID1}. Questions were divided in similar sets for other models based on their respective results.

\begin{table}[]
    \centering
    \begin{tabular}{c|cccc}
        \hline
        \textbf{Identity pair} & \textbf{C1} & \textbf{C2} & \textbf{C3} & \textbf{C4} \\
        \hline
        Asian, Indian & 6909 & 6820 & 164 & 149 \\
        helpful, Asian & 6846 & 6707 & 262 & 227 \\
        Asian, Yellow & 6888 & 6771 & 185 & 198 \\
        White, Black & 6866 & 6831 & 188 & 157 \\
        good, bad & 6834 & 6691 & 272 & 245 \\
        Indian, Brown & 6855 & 6781 & 203 & 203 \\
        good, sharp & 7035 & 6864 & 71 & 72 \\
        African, British & 6863 & 6787 & 197 & 195 \\
        helpful, White & 6808 & 6688 & 300 & 246 \\
        dumb, bad & 6824 & 6726 & 237 & 255 \\
        \hline
    \end{tabular}
    \caption{Number of questions in each subset for Llama 1B model}
    \label{tab:subsets}
\end{table}

\section{Additional Patching Results}
\label{sec:additional}
\begin{figure*}
    \centering
    \begin{subfigure}[t]{0.49\linewidth}
        \centering
        \includegraphics[width=\linewidth]{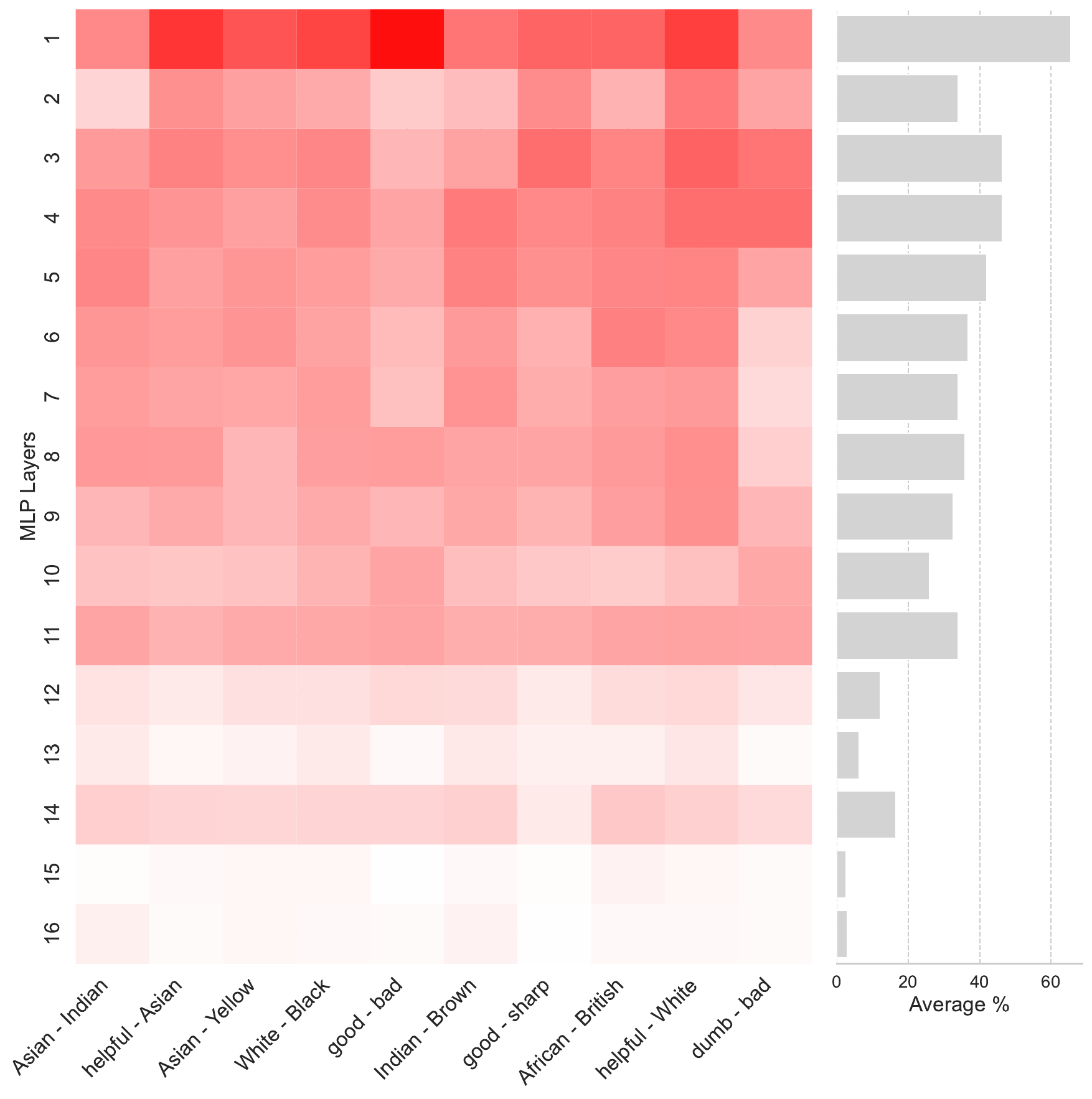}
        \caption*{MLP layer patching}
    \end{subfigure}
    \hfill
    \begin{subfigure}[t]{0.49\linewidth}
        \centering
        \includegraphics[width=\linewidth]{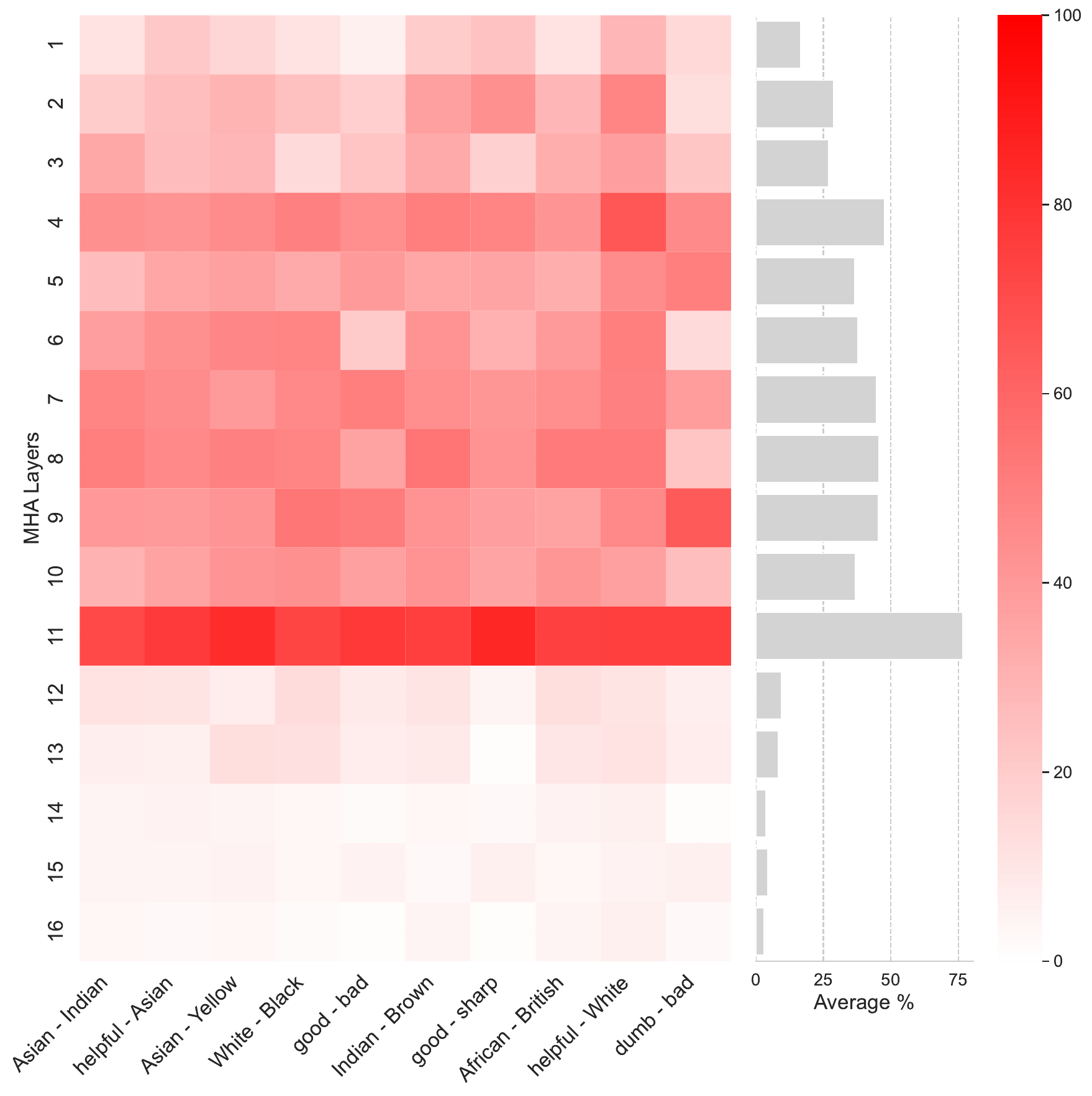}
        \caption*{MHA layer patching}
    \end{subfigure}
    \caption{Percentage of questions whose logit of correct token became maximum after patching MLP layer (left) and MHA layer (right).}
    \label{fig:is_max}
\end{figure*}

We also measured, for each identity pair \texttt{(ID1, ID2)}, the proportion of questions where \texttt{ID1} answered correctly and \texttt{ID2} did not, such that patching the activation of an MLP or MHA layer from \texttt{ID1}'s run into \texttt{ID2}'s run caused the correct token to receive the highest logit. Figure~\ref{fig:is_max} presents the results for both MLP and MHA layers, as well as their average across all identity pairs.

\begin{figure*}[t]
    \centering
    
    \begin{subfigure}[t]{0.49\linewidth}
        \centering
        \includegraphics[width=\linewidth]{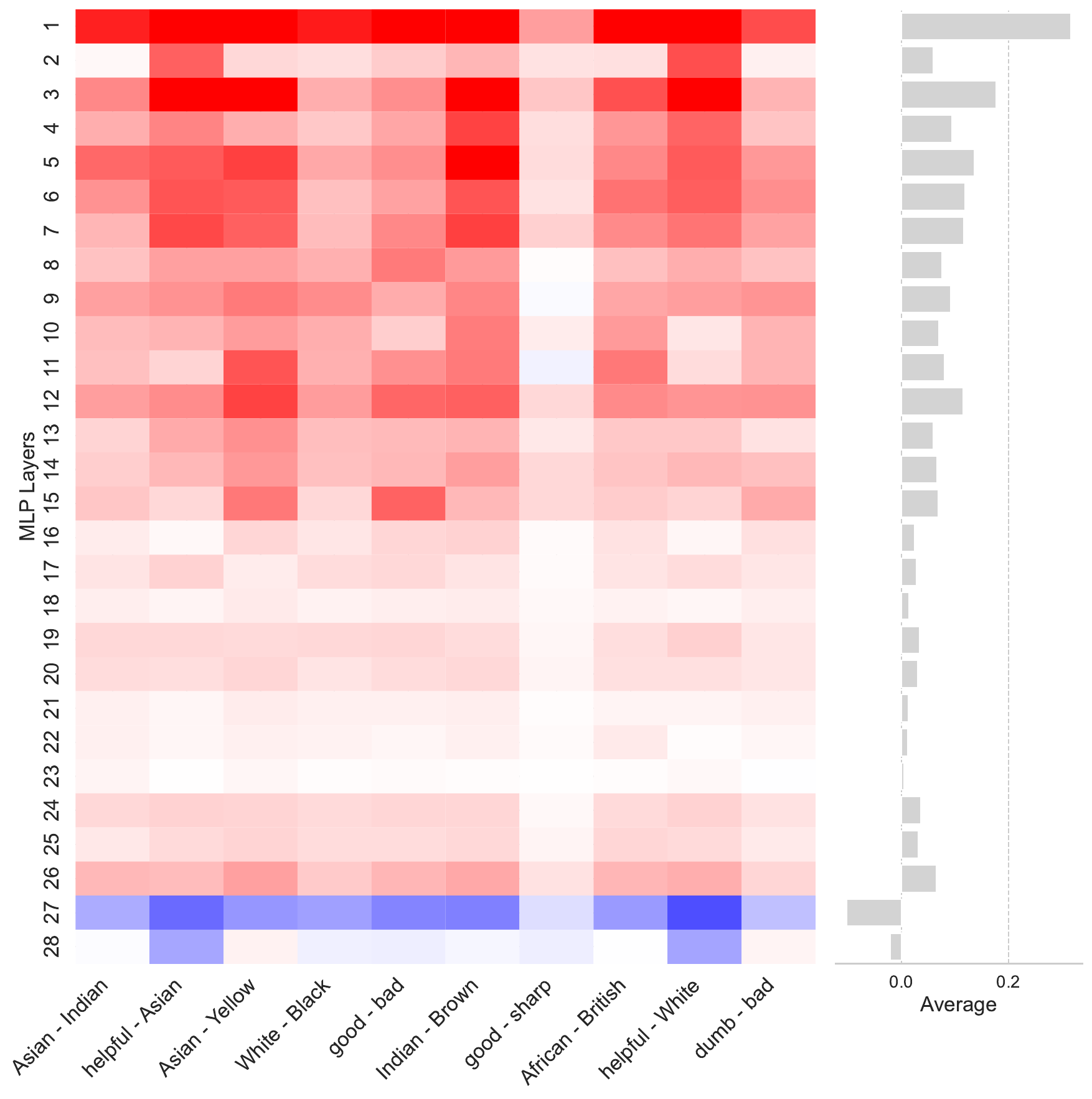}
        \caption*{MLP layer patching (Llama-3B)}
    \end{subfigure}
    \hfill
    \begin{subfigure}[t]{0.49\linewidth}
        \centering
        \includegraphics[width=\linewidth]{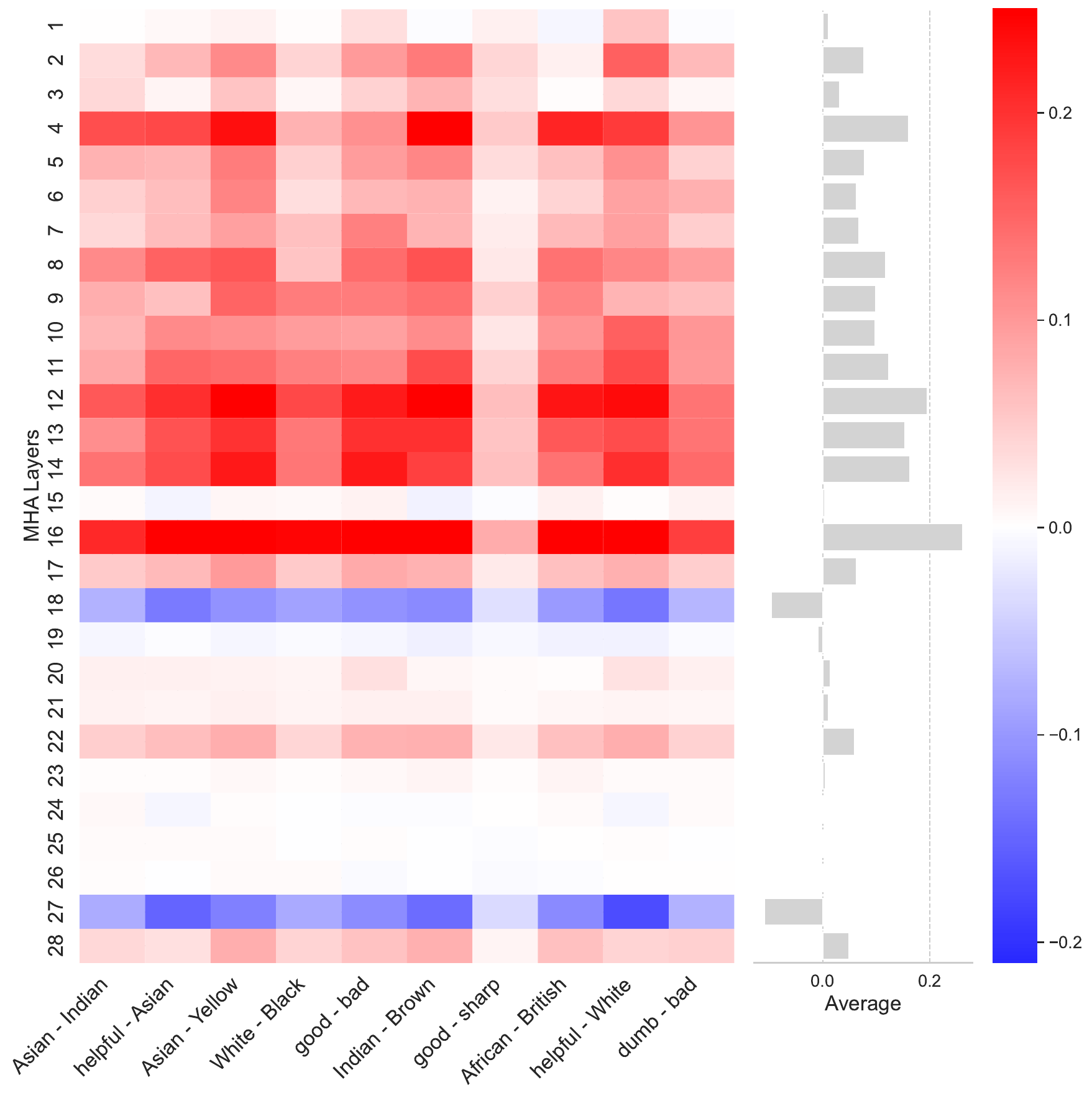}
        \caption*{MHA layer patching (Llama-3B)}
    \end{subfigure}

    \vspace{0.8em} 

    \begin{subfigure}[t]{0.49\linewidth}
        \centering
        \includegraphics[width=\linewidth]{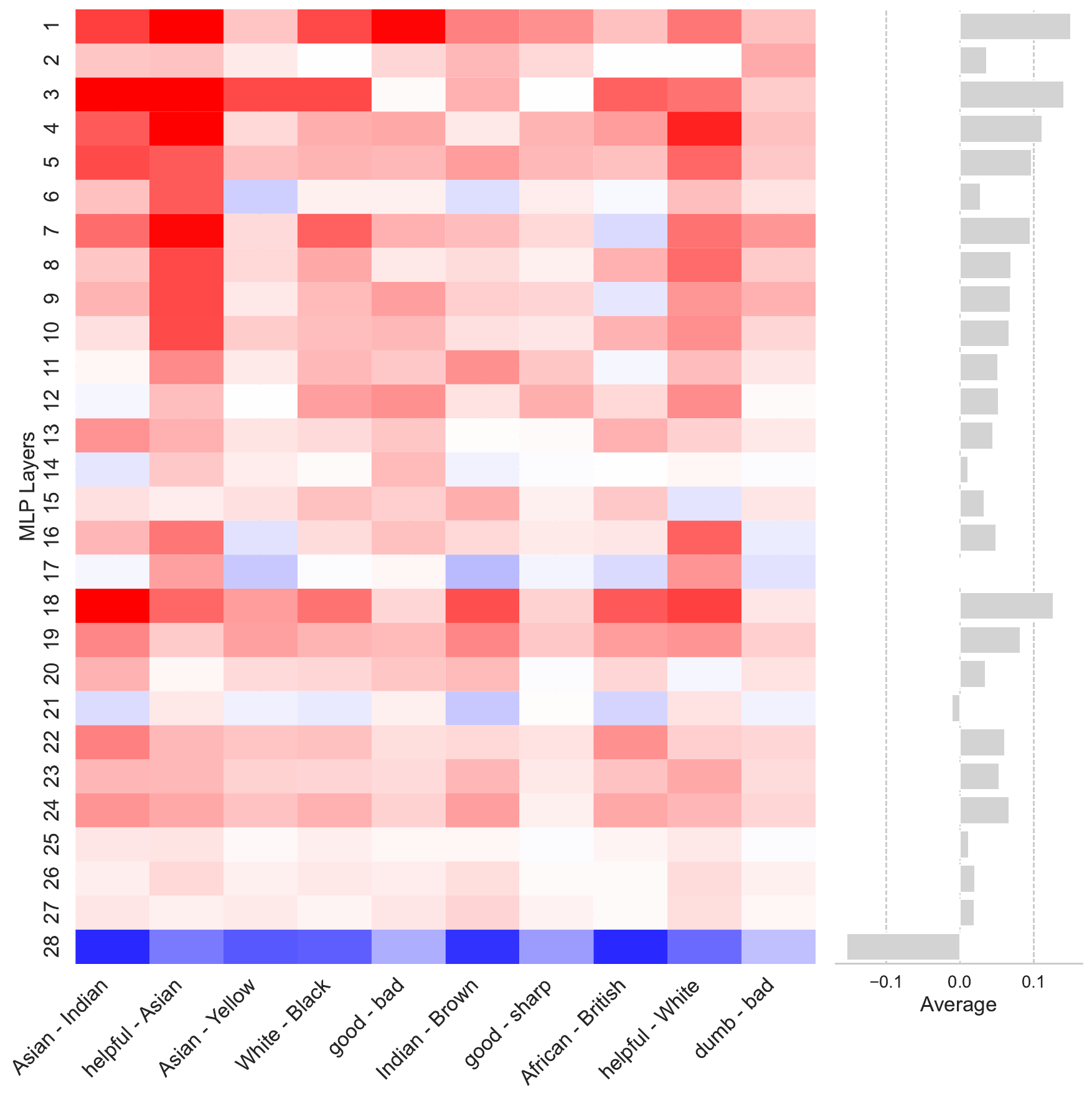}
        \caption*{MLP layer patching (Qwen-1.5B)}
    \end{subfigure}
    \hfill
    \begin{subfigure}[t]{0.49\linewidth}
        \centering
        \includegraphics[width=\linewidth]{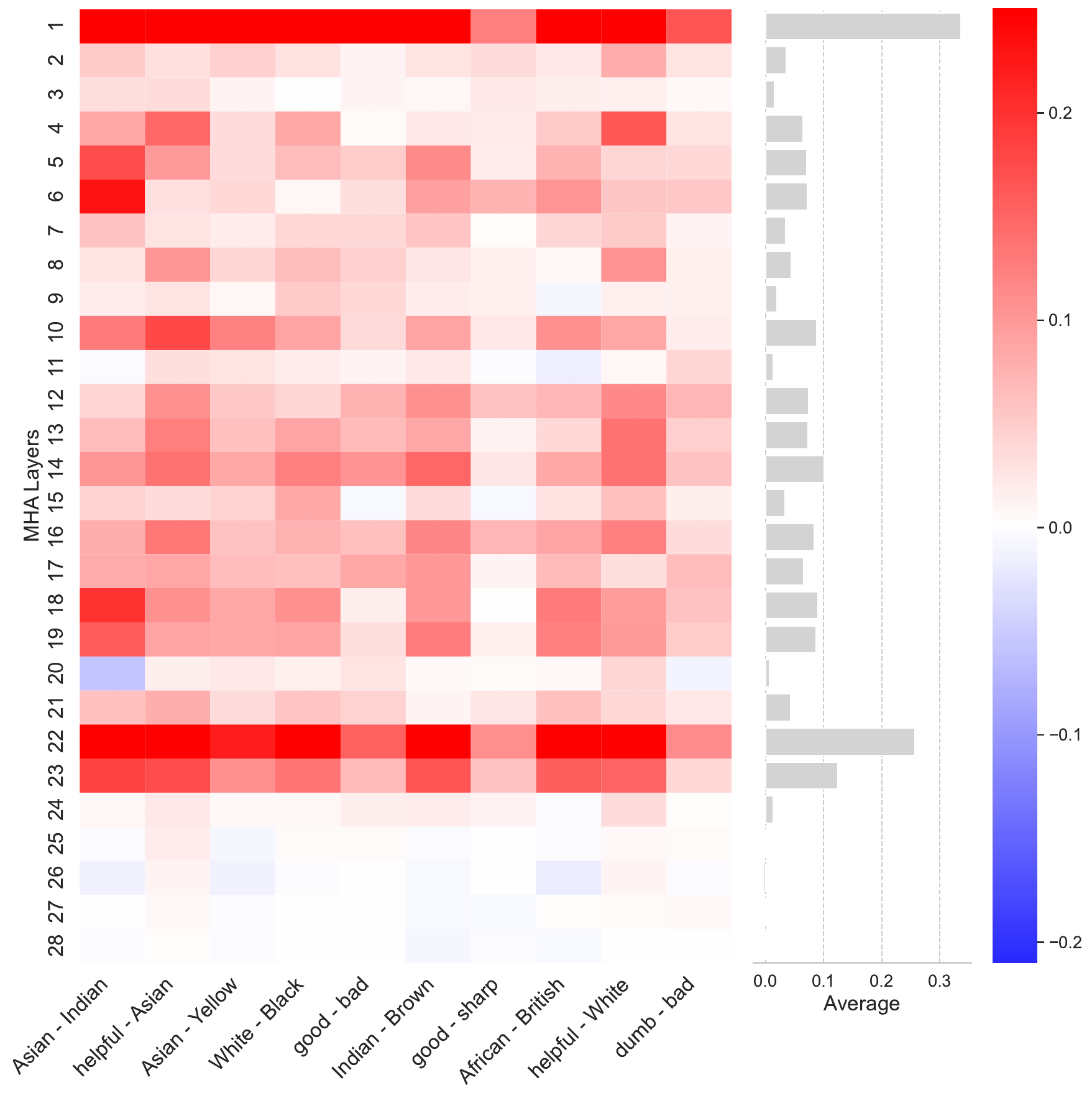}
        \caption*{MHA layer patching (Qwen-1.5B)}
    \end{subfigure}

    \caption{Relative logit difference ($\Delta_r$) when MLP layers (left) and MHA layers (right) are patched in Llama 3B (top row) and Qwen 1.5B (bottom row). Accompanying bar charts show the average $\Delta_r$ across identity pairs.}
    \label{fig:patching_results_models}
\end{figure*}

\section{Patching Results of other models}
\label{sec:patch_other}
We replicated the activation patching experiments on the MLP and MHA layers for the Llama 3B and Qwen 1.5B models. The results are shown in Figure~\ref{fig:patching_results_models}. We did not perform head-level patching on these models due to computational and time constraints. The results for the Llama 3B model are largely consistent with those of the Llama 1B model; however, the results for the Qwen 1.5B model show some anomalies, such as higher $\Delta_{r}$ values for the 18th MLP layer and the 1st MHA layer, which might be interesting to investigate further in future studies. Nevertheless, the main observations made in this work hold for all three models.


\section{Identity-token-position Patching Results}
\label{sec:id_patch}
\begin{figure*}[t]
    \centering
    \begin{subfigure}[t]{0.49\linewidth}
        \centering
        \includegraphics[width=\linewidth]{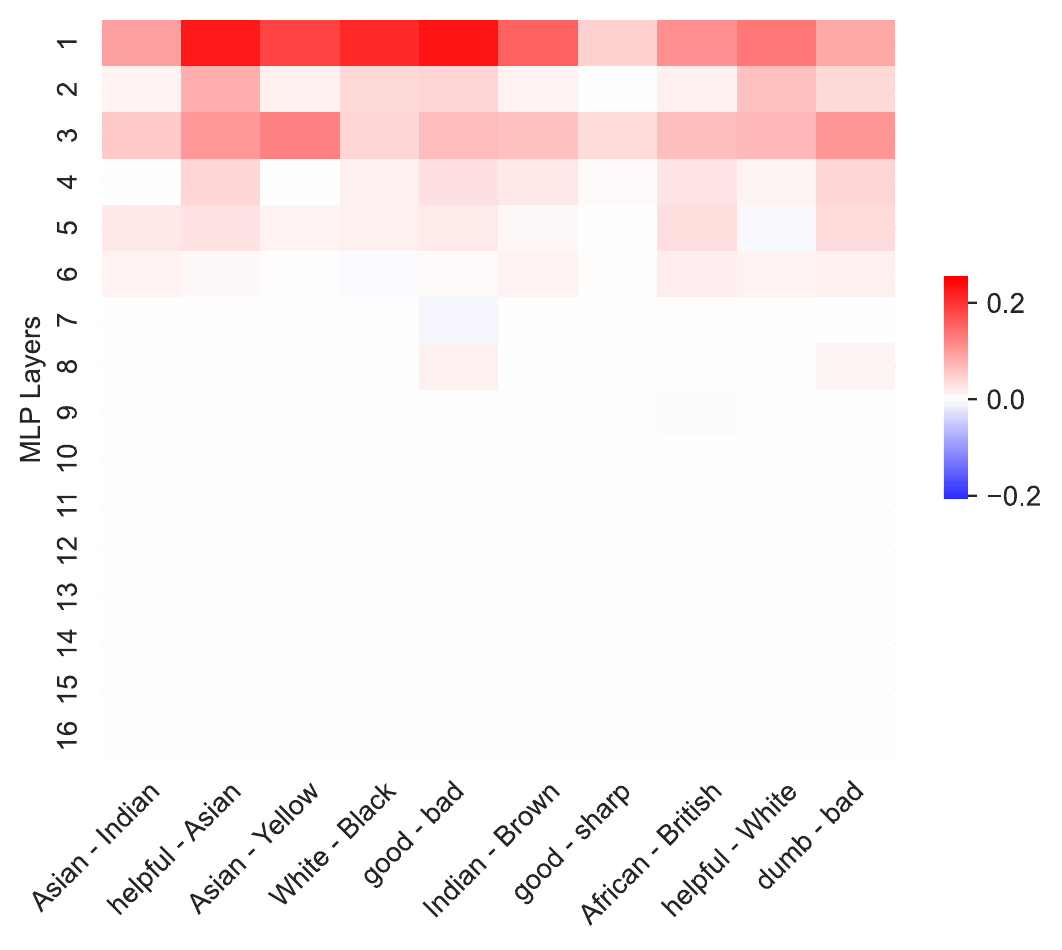}
        \caption{Relative logit difference ($\Delta_r$).}
        \label{fig:id_pos_mlp_relative}
    \end{subfigure}
    \hfill
    \begin{subfigure}[t]{0.49\linewidth}
        \centering
        \includegraphics[width=\linewidth]{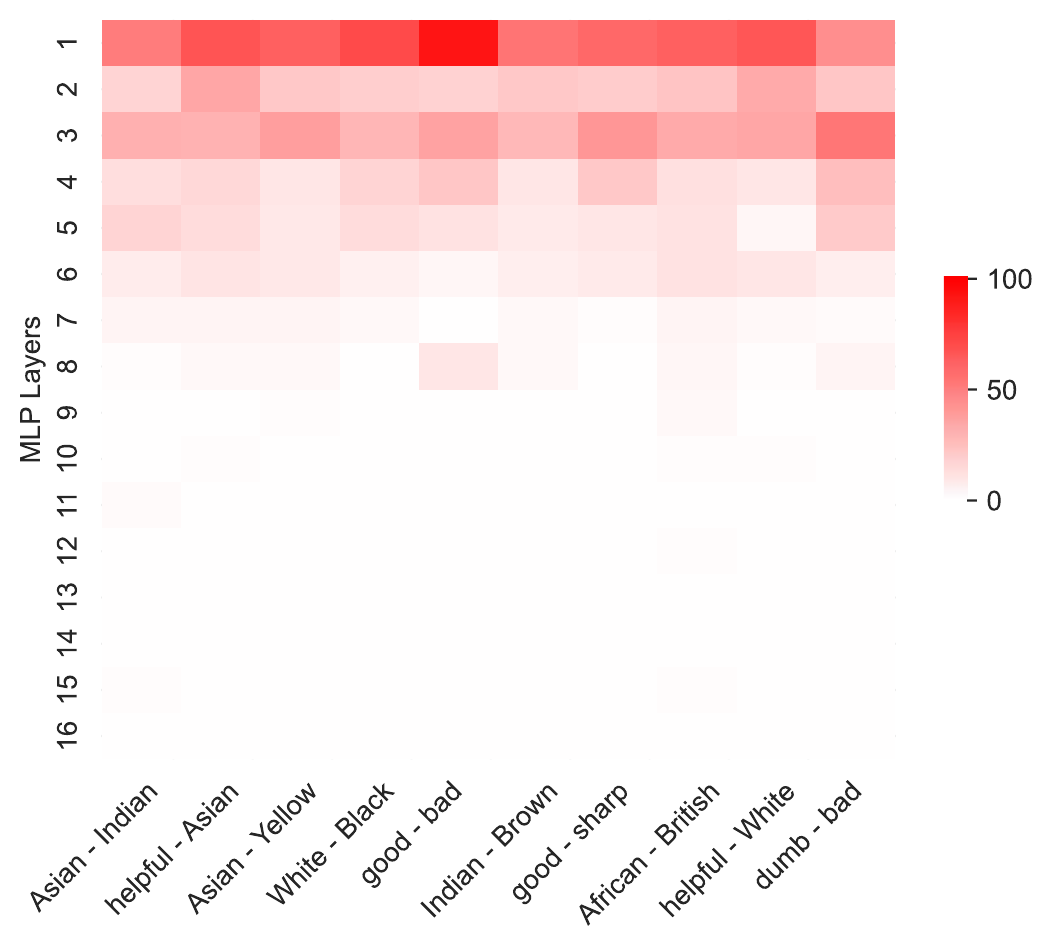}
        \caption{Percentage of questions whose logit of correct token became maximum.}
        \label{fig:id_pos_mlp_is_max}
    \end{subfigure}
    \caption{Comparison of MLP layer patching effects at the identity token position.}
    \label{fig:id_pos_mlp_side_by_side}
\end{figure*}

The relative change in the logit of the correct token, when only the identity token position's activation is patched for the MLP layer, is shown in Figure~\ref{fig:id_pos_mlp_relative}. Figure~\ref{fig:id_pos_mlp_is_max} shows the percentage of questions for which the correct token receives the highest logit.

\section{Attention Visualization}
\label{sec:attn_vis}
\begin{figure*}
    \centering
    \includegraphics[width=\linewidth]{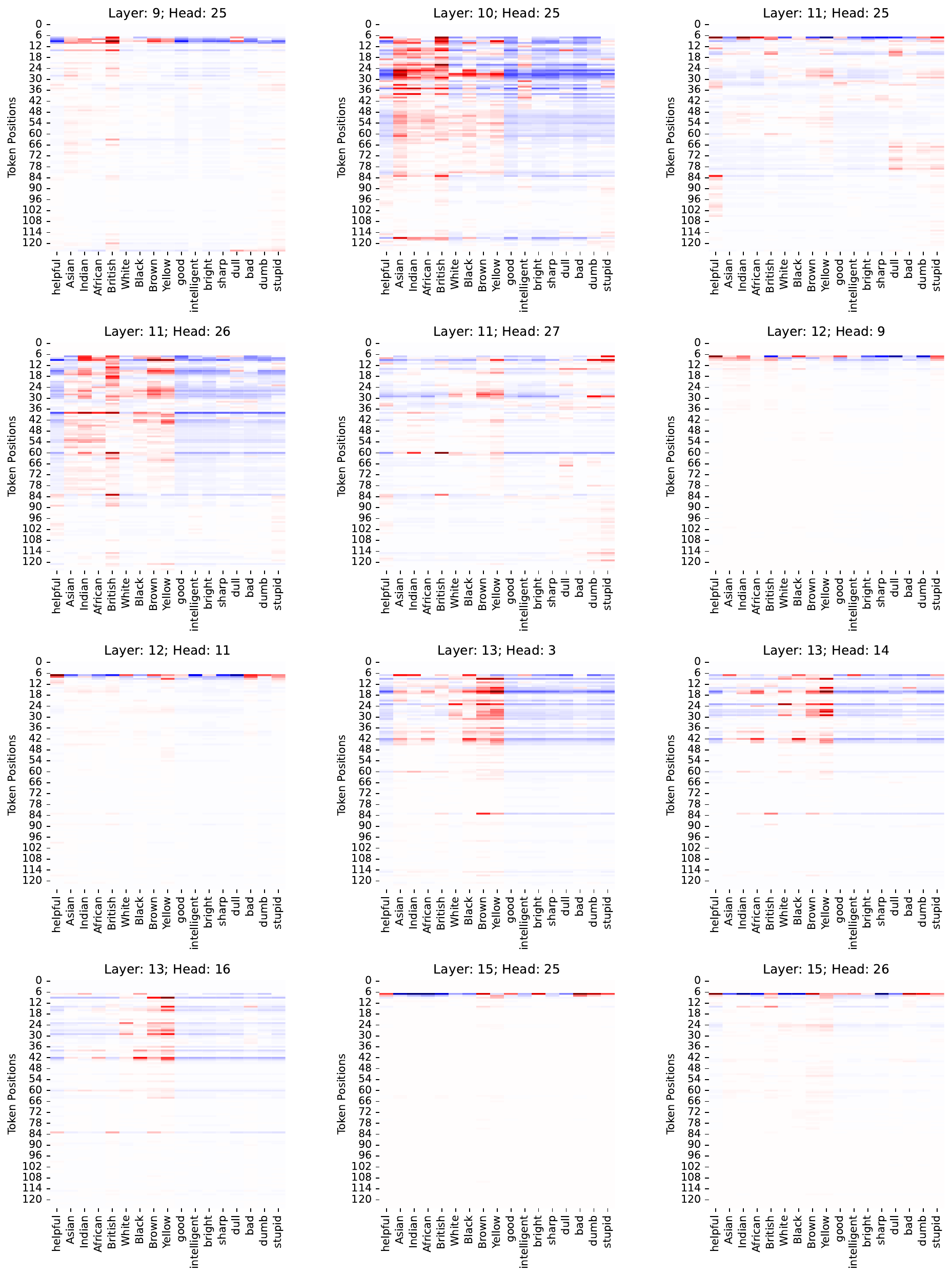}
    \caption{Relative value-weighted attention given by selected attention heads at identity token position. The prompt used is the first question from "Abstract Algebra" subject.}
    \label{fig:attention_vis}
\end{figure*}

Attention heads were selected based on high positive effect ($H_9^{25}$, $H_{10}^{25}$, $H_{11}^{25}$, $H_{11}^{26}$, $H_{11}^{27}$, $H_{12}^{9}$, $H_{13}^{14}$, $H_{15}^{25}$), and high negative effect ($H_{12}^{11}$, $H_{13}^{3}$, $H_{13}^{16}$, $H_{15}^{26}$). Figure~\ref{fig:attention_vis} shows the relative value-weighted attention given at the identity token position by selected attention heads, for a sample question from the dataset. Relative value-weighted attention is computed by subtracting the mean value-weighted attention across all identities from the value-weighted attention assigned to a given identity. This highlights the heads that pay disproportionately more attention to a specific identity or group of identities.

\section{Attention after patching}
\label{sec:attn_patch}

\begin{figure*}
    \centering
    \includegraphics[width=\linewidth]{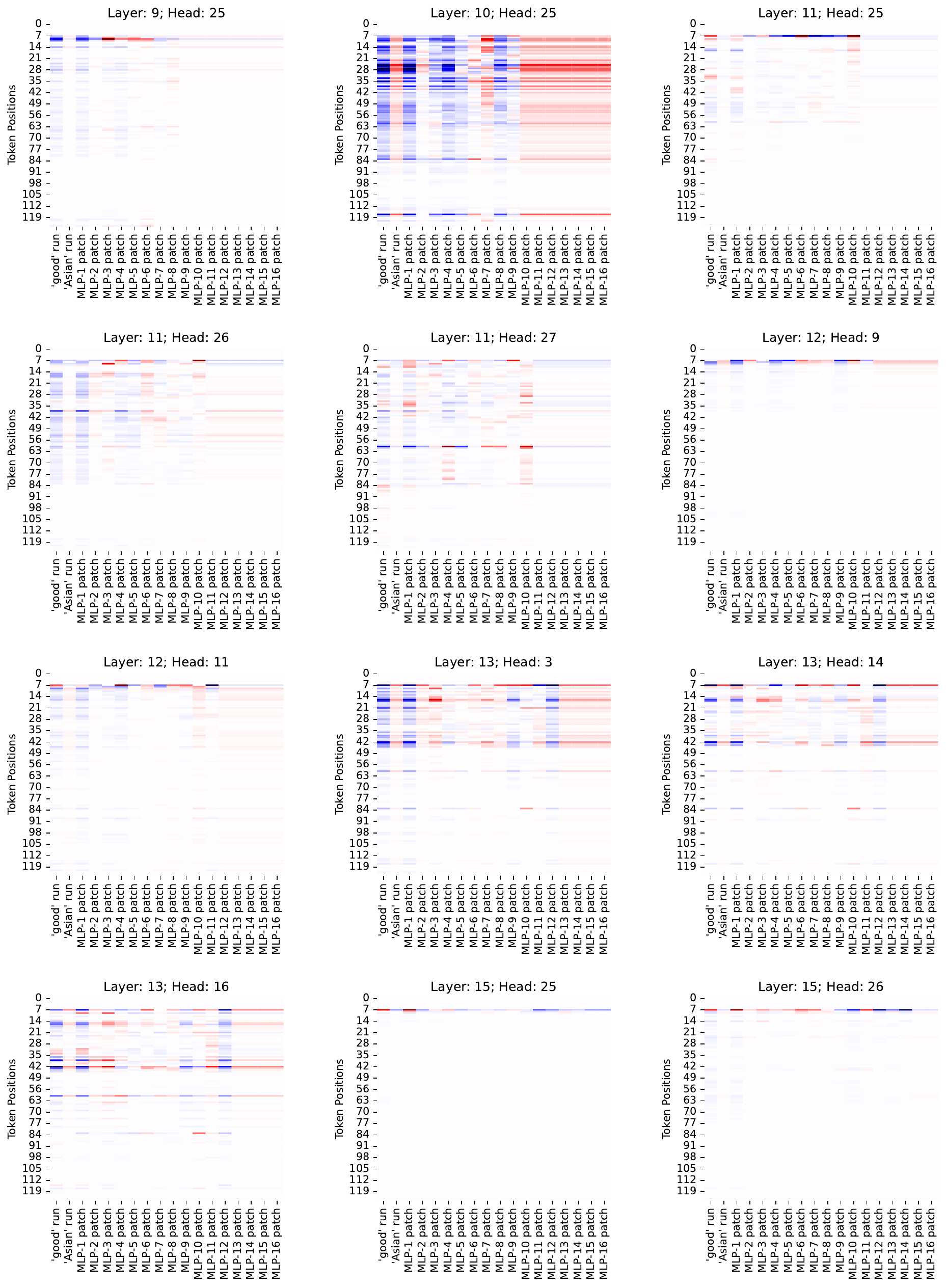}
    \caption{Relative value-weighted attention given by selected heads at identity token position after patching activation at identity token position for MLP layer. The prompt used is the first question from "Abstract Algebra" subject.}
    \label{fig:attention_vis_patched}
\end{figure*}

Figure~\ref{fig:attention_vis_patched} shows the change in value-weighted attention at the identity position when MLP layer activations are patched from the "good" run into the "Asian" run for a given question. The results indicate that racial heads (attention heads that allocated higher attention to racial-based identity tokens) assigns significantly less attention when the activations of early MLP layers are patched.

\end{document}